\documentclass[letterpaper, 10 pt, journal, twoside]{IEEEtran}  

\IEEEoverridecommandlockouts                              

\usepackage{amsmath} 
\usepackage{amssymb}  
\usepackage[utf8]{inputenc}
\usepackage[keeplastbox]{flushend}
\usepackage{graphicx}
\usepackage{cite}
\usepackage{siunitx}
\usepackage{algorithm}
\usepackage{algpseudocode}
\usepackage{afterpage}
\usepackage{gkmacros}
\usepackage{booktabs}
\usepackage{siunitx}
\usepackage{subfig}
\usepackage[nolist]{acronym} 
\usepackage{mathdots} 

\title{Multiple Hypothesis Semantic Mapping for Robust Data Association
\author{Lukas Bernreiter, Abel Gawel, Hannes Sommer$^{1}$, Juan Nieto, Roland Siegwart and Cesar Cadena}
\thanks{Manuscript received: February, 24, 2019; Revised May, 16, 2019;
Accepted June, 10, 2019.}
\thanks{This paper was recommended for publication by Editor Cyrill Stachniss upon
evaluation of the Associate Editor and Reviewers' comments.}
\thanks{This work was supported by the National Center of Competence in Research (NCCR) Robotics through the Swiss National Science Foundation and has received funding from the European Union’s Horizon 2020 research and innovation programme under grant agreement No 688652 and from the Swiss State Secretariat for Education, Research and Innovation (SERI) under contract number 15.0284.}
\thanks{All authors are with the Autonomous Systems Lab, ETH Zurich, Zurich 8092, Switzerland, {\tt \small \{berlukas, gawela, sommerh, nietoj, rsiegwart, cesarc\}@ethz.ch.}}
\thanks{$^1$ Additionally with Sevensense Robotics AG, Zurich.}%
\thanks{Digital Object Identifier (DOI): see top of this page.}
}

\acrodef{DP}{Dirichlet Process}
\acrodef{MHT}{Multiple Hypothesis Tracking}
\acrodef{MHt}[MHt]{Multiple Hypothesis Tree} 
\acrodef{RFS}{Random Finite Set}
\acrodef{PMF}{Probability Mass Function}
\acrodef{PDF}{Probability Density Function}
\acrodef{VO}{Visual Odometry}
\acrodef{LO}{Laser Odometry}
\acrodef{RMSE}{Root Mean Square Error}
\acrodef{UKF}{Unscented Kalman Filter}
\acrodef{JSD}{Jensen-Shannon divergence}
\acrodef{ML}{Maximum Likelihood}

\begin{document}

\maketitle

\begin{abstract}	
In this paper, we present a semantic mapping approach with multiple hypothesis tracking for data association. 
As semantic information has the potential to overcome ambiguity in measurements and place recognition, it forms an eminent modality for autonomous systems. 
This is particularly evident in urban scenarios with several similar looking surroundings. 
Nevertheless, it requires the handling of a non-Gaussian and discrete random variable coming from object detectors.
Previous methods facilitate semantic information for global localization and data association to reduce the instance ambiguity between the landmarks.
However, many of these approaches do not deal with the creation of complete globally consistent representations of the environment and typically do not scale well. 
We utilize multiple hypothesis trees to derive a probabilistic data association for semantic measurements by means of position, instance and class to create a semantic representation.
We propose an optimized mapping method and make use of a pose graph to derive a novel semantic SLAM solution. 
Furthermore, we show that semantic covisibility graphs allow for a precise place recognition in urban environments. 
We verify our approach using real-world outdoor dataset and demonstrate an average drift reduction of 33$\,$\% w.r.t. the raw odometry source. 
Moreover, our approach produces 55$\,$\% less hypotheses on average than a regular multiple hypotheses approach. 

\end{abstract}


\section{Introduction}
\label{sec:introduciton}
Semantic data is a reliable and ubiquitous flow of information in structured and non-structured environments. 
Especially for perception systems, semantically annotated data and higher reasoning about the underlying scene on top of purely geometric approaches have the potential to increase the robustness of the estimation~\cite{Cadena2016, ess2009segmentation}.
A reliable mapping is eminently important especially for autonomous, as well as, augmented reality systems since the recognition of the surrounding objects and the localization in a globally unknown environment are crucial factors there.

Traditional approaches for localization often rely on specific low-level visual features such as points and lines which are inherently ambiguous preventing the approach to scale well to large environments. 
In contrast, semantic information features a promising approach for many robotic applications by allowing more unique local and global descriptors for landmarks as well as potential viewpoint-invariance. 
Therefore, this constitutes a crucial factor for the measurement association to mapped landmarks and thus influences the quality of the localization. 
Moreover, semantics are very efficient at dealing with place recognition as they are less affected by seasonal or appearance changes as well as large drifts. 

In a conventional SLAM setting, the measurement noise is commonly relaxed to the continuous Gaussian case~\cite{Cadena2016} which however, does not apply to semantic variables. 
Uncertainties in the object detection such as class labels and object instances typically involve the handling of non-Gaussian discrete variables. 
How to properly handle such variables is still quite challenging and remains an open research question~\cite{blum2018modular}. 
\begin{figure}[!t]
   \centering
   \includegraphics[width=0.48\textwidth, trim={0cm, 0cm, 0cm, 0cm}, clip]{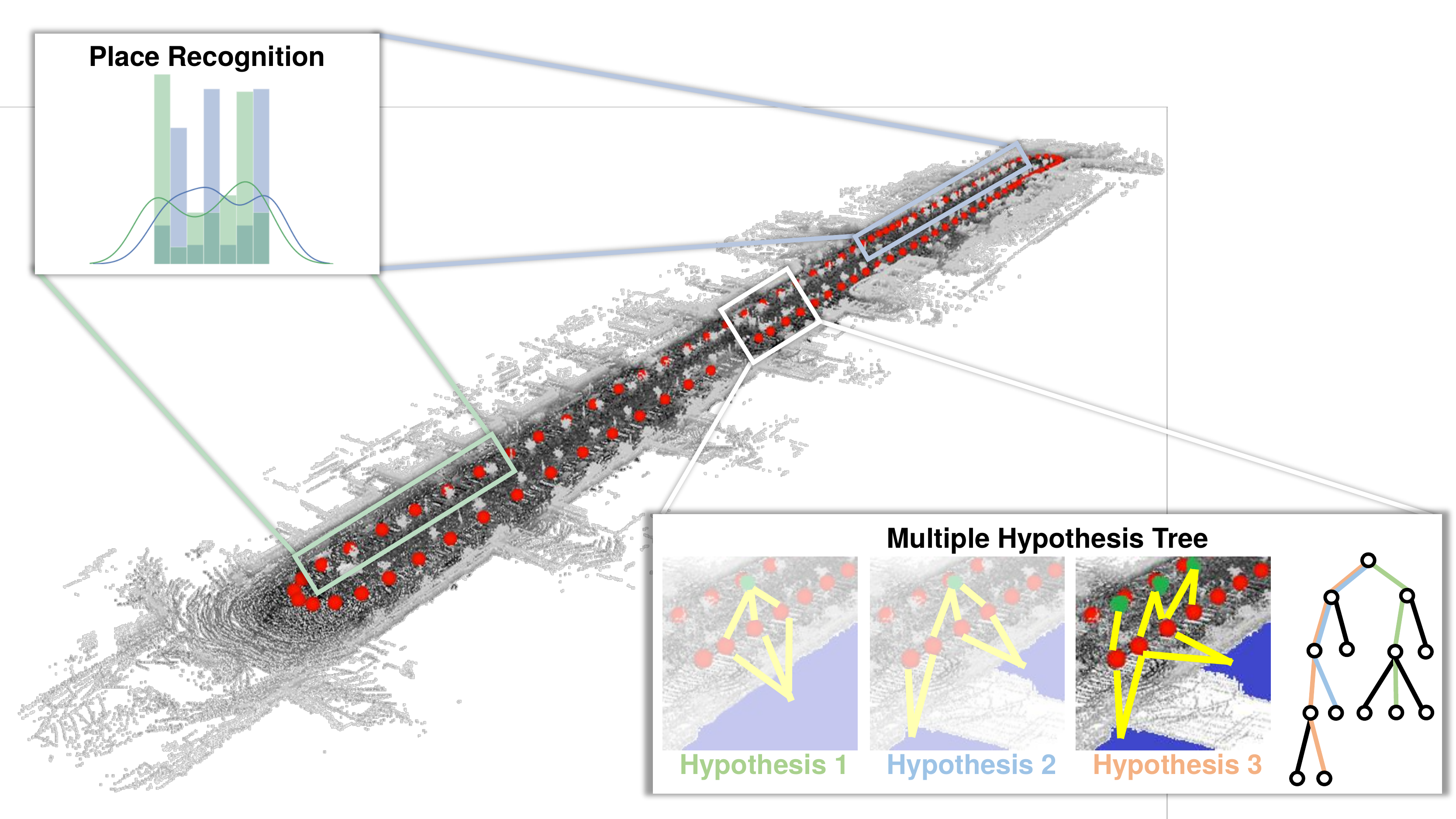}
   \caption{We propose a semantic SLAM system that maintains multiple hypotheses of the landmark locations structured in a hypothesis tree (bottom right image). Data association is done in a semantic framework to create new branches in the hypothesis tree. Furthermore, we perform a semantic place recognition method utilizing the object class distribution of a submap (top left image).}
   \label{pics:cover}
\end{figure}
Many existing semantic mapping approaches are primarily concerned with the creation of an indoor semantic representation with minor illumination and viewpoint changes~\cite{Kostavelis2015a}.
In contrast, realistic outdoor applications often come with severe changes of illumination and viewpoint.
This can hamper loop closure detection since drastic view-point changes might render scenes completely different when revisiting.
\begin{figure*}[!ht]
  \vspace{0.2cm}
  \centering
   \includegraphics[width=0.62\textwidth, trim={0.4cm, 5.8cm, 0.5cm, 6cm}, clip]{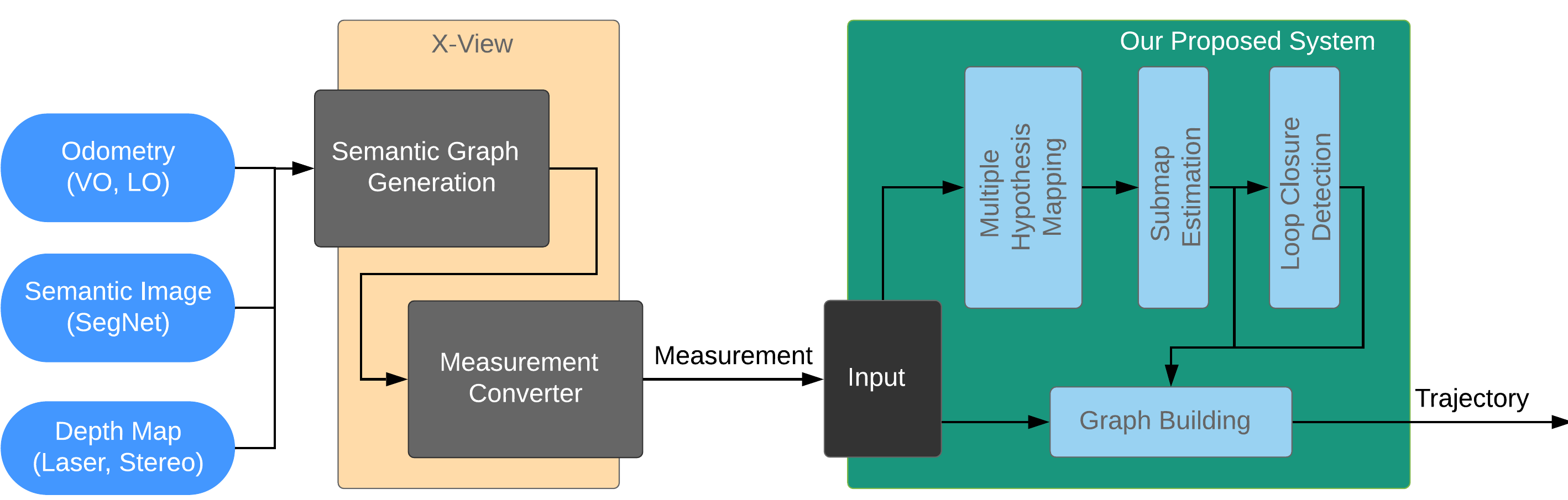}
   \caption{System overview of our proposed approach. We make use of our previous work X-View~\cite{Gawel2017a} by means of the semantic object extraction. Hereby the semantic objects are converted to measurements and used as input for our system. Afterwards we start simultaneously with the creation of the factor graph as well as with the multiple hypothesis mapping of the environment. Loop closures are detected and placed into the factor graph when submaps are completed.}
   \label{pics:mapping:overview}
\end{figure*}

Additionally, local descriptors for place recognition often rely on the bag-of-words paradigm~\cite{Labbe2013,Angeli2008} which can fail often in environments with repetitive features which commonly occur in urban environments leading to false loop closures.

Other semantic SLAM systems do not directly incorporate the semantic information into the estimation but rather use it to filter out bad classes such as cars or pedestrians beforehand~\cite{chen2017monocular}.

In this work, we aim to build a globally consistent semantic mapping formulation by improving the incorporation of discrete random variables in the map building and localization processes.
Throughout this paper, observations comprise the semantic class and position from static landmarks as well as the spatial relationship to other static landmarks.
We approach the measurement association problem by utilizing the semantic class of an object and deferring the decision on associations until the ambiguity is resolved.
In other words, the decision on the association is done at a time when more observations are available or a place is revisited allowing to correctly identify the instance label with a certain assurance.
This is motivated by the fact that in many cases the most likely association given only a few measurements does not necessarily need to be the correct one.

Furthermore, we derive a loop closure detection and verification algorithm operating directly on the level of semantic objects. 
Utilizing the class labels and the spatial relationships between the objects enables a robust recognition of places in urban environments.
An overview of our proposed system is given in figure~\ref{pics:mapping:overview}. 

The main contributions of this work are
\begin{itemize}
    \item Consistent multiple hypothesis mapping using an optimized \ac{MHT} approach.
    \item A \ac{DP}-based relaxed probabilistic Hungarian algorithm for viewpoint-invariance. 
    \item Semantic selection strategy to identify potential submaps for loop closures.
    \item Place recognition based on the semantic classes and the covisibility graphs.
    \item Incorporation of the proposed approach into a graph-based semantic SLAM pipeline and evaluation of the resulting system.
\end{itemize}

\subsection{Related Work}
In recent years, the advances to deep learning systems led to more reliable as well as practically usable object detectors~\cite{garcia2017review}. 
Consequently, this allowed SLAM systems to additionally include semantically rich information in order to improve their estimation~\cite{Salas-Moreno2013, Civera2011, hosseinzadeh2018structure}.
Recently, some research specifically addresses the problem of correctly assigning measurements to already known objects utilizing additional semantic information~\cite{Elfring2013, Wong2016}.
These systems, however, do not deal with the estimation of the camera's position, i.e. their application implies a static position of the camera and is often placed indoors.
Thus, they are not optimized for viewpoint-invariance, but rather emphasize on the probabilistic data association and the tracking of objects across multiple scenes.

Nevertheless, we make use of the close relationship to robotic mapping since target tracking is a special case of mapping.
Elfring et al. \cite{Elfring2013} presented a semantic anchoring framework using \ac{MHT}s \cite{Blackman2004} which defers the data association until the ambiguity between the instances is resolved. 
Generally, the \ac{MHT} enables accurate results but is inherently intractable with a large amount of objects and requires frequent optimizations \cite{Cox1994}. 
The work of Wong et al. \cite{Wong2016} presents an approach using the \ac{DP}s which yields estimation results comparable to the \ac{MHT} but with substantially less computational effort. 
Nevertheless, their proposed approach is not incremental and therefore, not directly applicable for the mapping of a robot's environment. 
Similar, in their previous work \cite{Wong2015} the authors propose a world modeling approach using dependent \ac{DP}s to accommodate for dynamic objects. 
In their proposed framework, the optimal measurement assignment is computed using the Hungarian method operating on negative log-likelihoods for the individual cases. 
Furthermore, Atanasov et al. \cite{Atanasov2016b} emphasizes on a novel derivation of the likelihood of \ac{RFS} models using the matrix permanent for localizing in a prior semantic map. 
Their system utilizes a probabilistic approach for data association which considers false positives in the measurements.

There is a vast literature on indoor semantic mapping available which however, does often not directly incorporate semantic information in a SLAM pipeline but rather uses the information for mapping and scene interpretation~\cite{nuchter2008towards}~\cite{pronobis2012large}.
The work that is most similar to ours is the work of Bowman et al.~\cite{Bowman2017} which proposed a semantic system which enables to directly facilitate semantic factors in their optimization framework.
Despite using probabilistic formulation for the data association, their approach inherently neglects false positives and false negatives, and lacks including a prior on the assignments. 
Moreover, they limited the possible classes in their mapping so that only cars were enabled in their outdoor experiments.
This greatly reduces the complexity in outdoor scenarios with semantically rich information and further, is not a reliable source for place recognition. 
Another direction of research is to represent landmarks as quadrics to capture additional information such as size and orientation~\cite{nicholson2018quadricslam,hosseinzadeh2018structure}.
However, they either assume that the measurement association is given~\cite{nicholson2018quadricslam} or utilize the semantic labels for a hard association using a nearest neighbor search~\cite{hosseinzadeh2018structure}.
Thus, their work does not include any probabilistic inference for the association and does not consider false positives.

    

Our previous work by Gawel et al.~\cite{Gawel2017a} represents the environment using semantic graphs and performs global localization by matching query graphs of the current location with a global graph.
The query graphs however, are not used in a data association framework and thus, landmarks could potentially be duplicated. 
This system does neither deal with map management and optimization nor with drift reduction for globally consistent mapping.
Our semantic SLAM system does not require any prior of the object shapes and comprises a soft probabilistic data association for semantic measurements.
To the best of our knowledge, a complete semantic SLAM system comprising the aforementioned approaches has not been reported in literature before. 

In the remaining part of this paper we will start deriving a semantic mapping approach (section ~\ref{sec:mapping}) and a concrete algorithm for localization (section~\ref{sec:localization}).
The presented work is evaluated in chapter \ref{sec:evaluation}. 
Finally, chapter \ref{sec:conclusion} concludes this work and gives further research directions. 

\section{Semantic Multiple-Hypotheses Mapping}
\label{sec:mapping}
When performing SLAM, measurement noise typically leads to drift and inconsistent maps -- in particular when measurements get wrongly associated to landmarks.
We approach this problem by introducing locally optimized submaps.
Each submap maintains an individual \ac{MHt} and propagates a first-moment estimate to proximate submaps. 
Specifically, for each submap we want to maximize the posterior distribution, $f(\mat{\Theta}_t |\mat{Z}_t)$,  of the associations $\mat{\Theta}_t$ of all measurements $\mat{Z}_t$ received till the time step $t$ which is proportional to%
\footnote{Vectors are underlined and matrices are written with bold capital letters.}%
\begin{align}
f(\vec{z}_t| \mat{\Theta}_t, \mat{Z}_{t-1}) f(\vec{\theta}_t | \mat{\Theta}_{t-1}, \mat{Z}_{t-1}) f(\mat{\Theta}_{t-1} | \mat{Z}_{t-1}) \label{eq:hyp_posterior}.
\end{align}
Here, the set of $N$ measurement associations at time step $t$ is represented by $\vec{\theta}_t=[{\theta}^1_t...{\theta}^N_t]$. 
The first factor, $f(\vec{z}_t| \mat{\Theta}_t, \mat{Z}_{t-1})$, in \eqref{eq:hyp_posterior} represents the distribution of all measurements at time $t$ for which we assume conditional independence of the individual measurements such that it equals to
\begin{align}
	\prod_{i=1}^{n} f (\vec{z}_t^i | \theta^i_t = l, \mat{Z}_{t-1})%
    = p_s(c^i_t) \prod_{i=1}^{n} p(c^i_t| \gamma_{t-1}^{l}) f(\vec{p}_t^i | \vec{\pi}_{t-1}^{l}),
    \label{eq:meas_likelihood}
\end{align}
where $\vec{z}_t^i$ denotes the attributes of the $i$th semantic measurement at time $t$, $\theta^i_t$ is the index of the landmark, $l$, this measurement is associated with, and $\vec{\pi}_{t-1}^{l}$, $\gamma_{t-1}^{l}$ the assigned landmark's position and class estimated at time $t-1$.
Furthermore, a semantic measurement, $\vec{z}^i_t$, is split into its position, $\vec{p}_t^i$, and class, $c_t^i$, component,
whose \ac{PMF}, $p_s$, is a prior assumption based on how well the classes fit into the current environment.
For the $i$th class measurement, $c^i_t$, we assume $p(c^i_t | \theta^i_t=l, \gamma_{t-1}^{l}):=\delta_{c^i_t,\gamma^{l}_{t-1}}$, where $\delta$ denotes the Kronecker delta.
\def\mnoise{\vec{\nu}}
For $\vec{p}_t^i$, we assume the following form of a stochastic measurement model
\begin{equation*}
	f(\vec{p}_t^i|\theta^i_t=l, \vec{\pi}_{t-1}^{l}) %
	\!=\!f_{\mnoise}(\vec{p}_t^i\!\! - \vec{\pi}_{t-1}^{l}),
\end{equation*}
where $f_{\mnoise}$ denotes the \ac{PDF} of the additive position measurement noise, $\mnoise_t^i$, which we model as a zero-mean Gaussian distribution with covariance $\mat{\Sigma}^z$.
%
For practical stability, we use an \ac{UKF}~\cite{Wan2000} for the estimation of $\pi^l_t$.
\begin{figure}[!t]
   \vspace{0.2cm}
   \centering
   \includegraphics[width=0.35\textwidth, trim={5cm, 9.4cm, 4cm, 7.4cm}, clip]{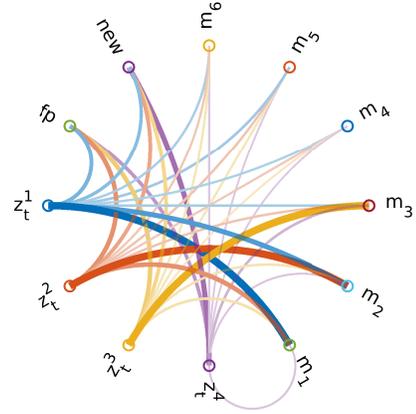}
   \caption{Likelihood of assigning a measurement to a specific scenario. Each measurement $z^i_t$, at time $t$, could be assigned to any of the existing landmarks  $m_{\{1..6\}}$, represent a new landmark (new) or a false positive (fp). 
   The thickness and opacity denotes how likely the association is given a certain example of measurements and landmarks.}
   \label{pics:mapping:assoc}
\end{figure}
The second factor, $f(\vec{\theta}_t | \mat{\Theta}_{t-1}, \mat{Z}_{t-1})$, in \eqref{eq:hyp_posterior} is the assignment prior and is calculated using the well-known equation~\cite{Elfring2013,Cox1994}
\begin{equation}~\label{eq:assignmentPrior}
f(\vec{\theta}_t | \mat{\Theta}_{t-1}, \mat{Z}_{t-1}) = \frac{N_t^n!N_t^f!}{N_t^m!} p_n(N_t^n) p_f(N_t^f),
\end{equation}
where $N_t^n$ denotes the number of new measurements, $N_t^f$ the number of false positives identified by the Hungarian method and $N_t^m$ the total number of measurements at time step $t$.  
The functions $p_n$ and $p_f$ are prior \acp{PMF} over the number of new measurements and false positives, respectively.
Typically  both are chosen as Poisson \acp{PMF} with a specific spatial density $\lambda$ and a volume $V$~\cite{Bar-Shalom2007}, e.g.
\begin{equation*}
p_n(N) = \exp\left(\lambda_nV\right)\frac{(\lambda_nV_n)^{N}}{N!}.
\end{equation*}
Each branch in the \ac{MHt} comprises a different set of associations $\mat{\Theta}_t$.
Utilizing \eqref{eq:meas_likelihood} and \eqref{eq:assignmentPrior} together with the previous posterior distribution $f(\mat{\Theta}_{t-1} | \mat{Z}_{t-1})$ we can evaluate these branches using \eqref{eq:hyp_posterior}.
\subsection{Probabilistic Measurement Association}
Finding the correspondence $\vec{\theta}_t$ between measurements and mapped object can be challenging since the current set of measurements often does not allow deriving a correct assignment. 
Fortunately, this problem can be considered as a weighted combinatorial assignment problem for which the Hungarian algorithm~\cite{Jonker1986, Wong2015} is well known.
Figure~\ref{pics:mapping:assoc} illustrates the probabilistic combinatorial assignment problem.
To find the most likely assignment we utilize a stochastic association algorithm based on the \ac{DP}.
\acp{DP} are a good choice for modeling the probability of seeing new and re-observing already mapped landmarks~\cite{Ranganathan2006}.

The likelihood of the associations of new measurements $\vec{z}_t$ at time $t$ with landmarks, $\vec{\theta}_t$, is expressed by $f(\vec{\theta}_t | \vec{z}_t, \mat{\Theta}_{t-1}, \mat{Z}_{t-1})$. 
We assume that at each time step a landmark in the scene can at most generate one observation.
Inspired by the dependent \ac{DP} formulation in \cite{Wong2015}, we differentiate four cases: (i) landmarks that have already been seen in the current submap, (ii) landmarks seen in previous submaps, (iii) new landmarks and (iv) false positives.  
The likelihood for the association of a measurement $\vec{z}^i_t$ with an existing landmark $k$ of the same class in the current submap is modeled as 
\begin{equation}
    f(\theta^i_t = k | \vec{z}^i_t, \mat{\Theta}_{t-1}, \mat{Z}_{t-1}) = \exp(N^k_t) f_\mnoise{}(\vec{p}^{i}_t - \vec{\pi}^{k}_t) .
\end{equation}
Here, the scalar $N^k_{t}$ denotes the number of assignments to the landmark $k$. 
Despite the fact that we only deal with static objects, a landmark, $l$, which was seen in a previous submap at time, $\tau$, is modeled using a transitional density, $T$, i.e.
\begin{equation}\label{eq:likelihood_previous_landmark}
    f(\theta^i_t = l | \vec{z}^i_t, \mat{\Theta}_{t-1}, \mat{Z}_{t-1}) =\!\! \int \!\! f_\mnoise{}(\vec{p}^{i}_t\!\! - \vec{x}) T(\vec{x}, \vec{\pi}_\tau^{l}) \, d\vec{x} \, .
\end{equation}
The transitional density, $T$, depends on the semantic class of the object and is used to accommodate for the unknown shape of the landmarks.
Since we take the centroid of the segmented objects as input, we employ two approaches for the choice of $T$ to compensate for large measurement noise. 
Objects such as poles and trees are modeled using a Dirac-$\delta$ distribution: $T(\vec{x} | \vec{\pi}_\tau^{l}) = \delta(\vec{x} - \vec{\pi}_\tau^{l})$, reducing the right hand side of \eqref{eq:likelihood_previous_landmark} to $f_\mnoise{}(\vec{p}^{i}_t - \vec{\pi}_{\tau}^{l})$, the measurement distribution of $\vec{z}^i_t$ given the last seen position of $l$.
The transition of objects like buildings and fences is modeled using a Gaussian distribution with covariance $\mat{\Sigma}^\alpha$ resulting in 
\begin{align*}
    \int f_{\mnoise} & (\vec{x} - \vec{p}_t^{i} ) \mathcal{N}(\vec{x}; \vec{\pi}_\tau^{l}, \mat{\Sigma}^\alpha) \, d\vec{x} \\
    &= \mathcal{F}^{-1}\{\mathcal{F}\{\mathcal{N}(0, \mat{\Sigma}^z)\}\cdot\mathcal{F}\{\mathcal{N}(\vec{\pi}_\tau^{l}, \mat{\Sigma}^\alpha)\} \}(\vec{p}_t^{i})\\
    &= \mathcal{N}(\vec{p}_t^{i}; \vec{\pi}_\tau^{l}, \mat{\Sigma}^z+\mat{\Sigma}^\alpha),
\end{align*}
where $\mathcal{F}$ is the Fourier transform and $\mat{\Sigma}^\alpha$ is a transitional covariance depending on the object's class $c$.
The likelihood of assigning a new landmark, $l$, to the $i$th observation at time $t$, is approximated by the uniform distribution in $\vec{z}^i_t$ over the volume of the map~\cite{Wong2015}, $\mathcal{M}$, i.e.
\begin{align*}
    f({\theta}_t^i = l | \vec{z}_t^i, \mat{\Theta}_{t-1}, \mat{Z}_{t-1}) &= \alpha \int f_\mnoise{}(\vec{p}_t^i - \vec{x}) H_{\text{DP}}(\vec{x})  \, d\vec{x} \\
    &\approx \alpha U_{|\mathcal{M}|} \, ,
\end{align*}
where $H_{\text{DP}}$ is the base distribution of the \ac{DP}. 
Generally, false positives occur due to clutter in the images and are essentially detected objects which do not physically exist in the environment. 
Having such cases in the map may result in improper assignments of future measurements. 
False positives have the likelihood $f(\theta_t^i = 0 | \vec{z}_t^i, \mat{\Theta}_{t-1}, \mat{Z}_{k-1})$, i.e. the likelihood of the measurement $i$ being an observation of the \emph{false} landmark, $0$.
False positives, are assumed to occur at a fixed rate $\rho$, i.e.
\begin{align*}\label{eq:mhm:fp}
  f(&\theta_t^i = 0 | \vec{z}_t^i, \mat{\Theta}_{t-1}, \mat{Z}_{t-1}) \\ 
  & \propto \Big( \prod_j f(\vec{z}_t^i | \theta_t^i = j, \gamma_{t-1}^j, \vec{\pi}_{t-1}^j) \Big)^{-1} \cdot
  \begin{cases}
  	\rho N^0_t & N^0_t > 0 \\
    \rho \alpha & N^0_t = 0
  \end{cases},
\end{align*}
where $N^0_t$ denotes the number of occurred false positives until time step $t$, and $\alpha$ is the concentration parameter of the \ac{DP}. 
\begin{figure}[!t]
   \centering
   \vspace{0.2cm}
   \includegraphics[width=0.35\textwidth, trim={12cm, 4.5cm, 12cm, 4.0cm}, clip]{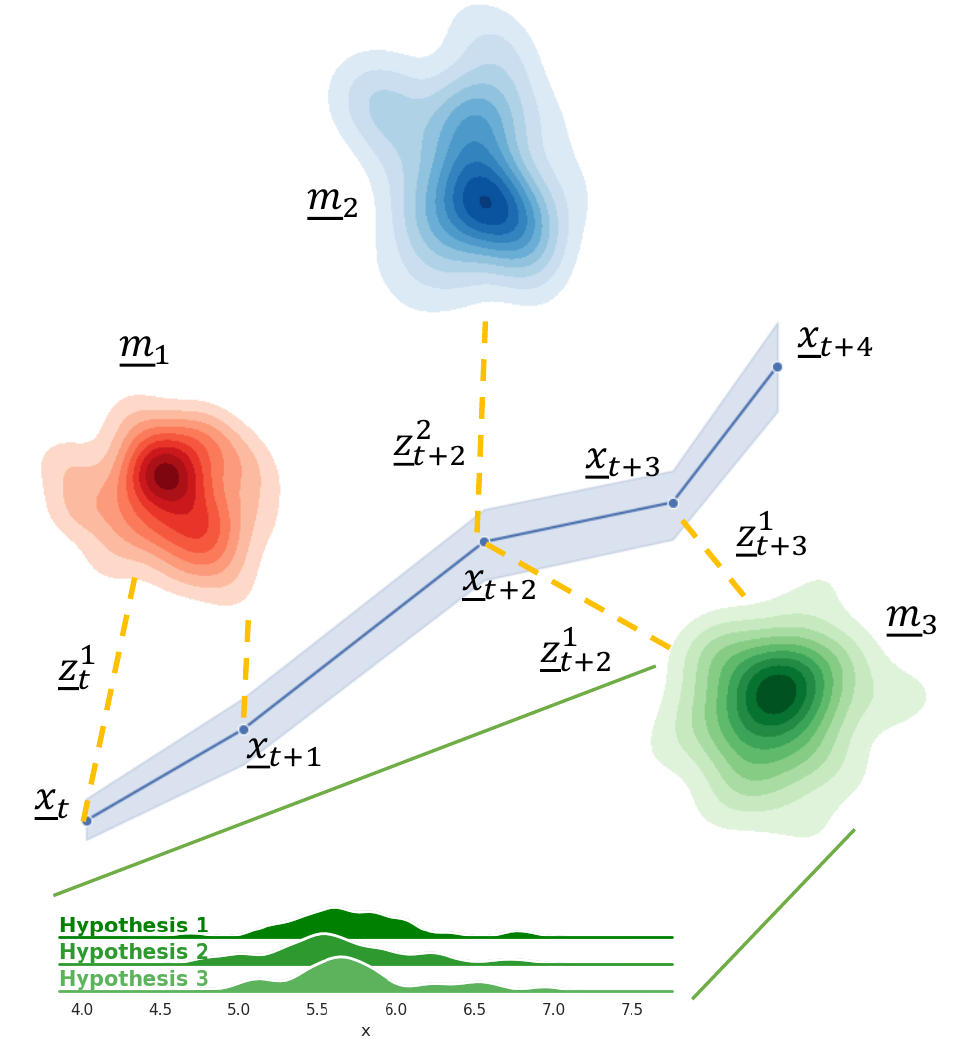}
   \caption{Illustration of the Gaussian mixture landmarks created by the fusion of the weighted hypotheses. }
   \label{pics:traj_with_GM}
\end{figure}
The aforementioned four cases will be used as an input to the Hungarian algorithm yielding an optimal assignment for each measurement as well as landmark. 
Based on this initial assignment, the optimal branch of the \ac{MHt} will be formed. 
In case the assignment is not distinct enough, new branches in the \ac{MHt} are generated by re-running the Hungarian algorithm without the previous optimal assignment. 
Since small \acp{MHt} are generally better for computational performance, we only create branches for associations that are reasonable. 
\begin{figure*}[!ht]
   \vspace{0.2cm}
   \centering
   \includegraphics[width=0.9\textwidth, trim={0.5cm, 5.5cm, 1cm, 5.1cm}, clip]{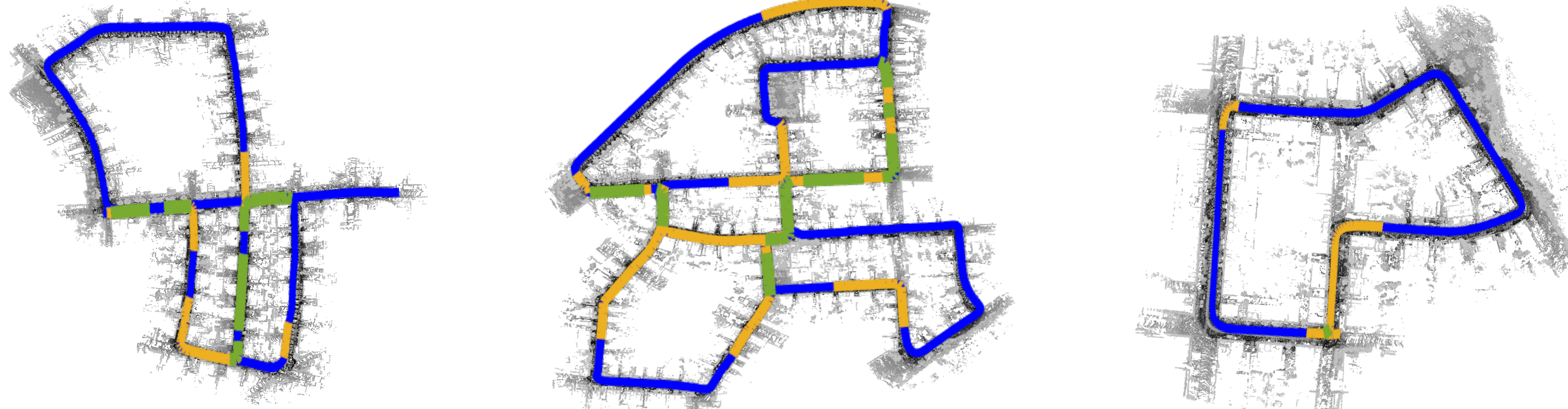}
   \caption{Illustration of the trajectories of the KITTI sequence 05, sequence 00, sequence 07 together with a laser map. Regions in blue denote the estimated trajectory, orange regions are submaps which were checked for loop closures and green areas show performed loop closures. We gain additional efficiency by only checking a subset of the submaps for loop closure.}
   \label{pics:evaluation:maps}
\end{figure*}
\subsection{Optimized Resampling of Hypotheses}
Each hypothesis is weighted using their measurement likelihood~\eqref{eq:meas_likelihood} and assignment prior term~\eqref{eq:assignmentPrior}.
At each time step, the existing hypotheses are reweighted and eventually resampled by a systematic resampling technique~\cite{Ristic2004}. 
In general, this fuses the current knowledge in the hypothesis set and eliminates the hypotheses which have a low weight and preserves hypotheses with a good weight. 

A crucial factor is when to decide that resampling should be performed on the hypothesis tree.
In this case, it is common to use selective resampling~\cite{Grisetti2005} based on the calculation of the effective sample size which essentially captures the diversity of the hypothesis set. 
Consequently, resampling is only performed when the effective sample size exceeds a certain threshold. 
Furthermore, many particle filter implementations only consider a fixed particle size.
However, it is desired that the number of particles is high for a high state uncertainty, and low when the uncertainty is low. 
Fox~\cite{Fox2003} introduced a variable sampling algorithm based on the KLD distance for particle filters. 
During each iteration of the resampling procedure, the number of hypotheses is dynamically bounded by $n$ using
\begin{equation}\label{eq:mhm:kldResample}
  n = \frac{k-1}{2\epsilon}\left(1-\frac{2}{9(k-1)}+\sqrt{\frac{2}{9(k-1)}}z_{1-\delta}\right),
\end{equation} 
where $k$ is the current number of resampled hypotheses and $z_{1-\delta}$ is the upper $1-\delta$ quantile of a normal distribution which models how probable the approximation of the true sample size is~\cite{Fox2003}.
The value $n$ is dynamically calculated at each step of the resampling until the number of resampled hypotheses is greater than $n$. 
Nevertheless, we bound the maximum number of resampled hypotheses to avoid drastic changes.
\section{Semantic Localization}
\label{sec:localization}
\begin{figure*}[!t]
   \vspace{0.2cm}
   \centering
   \includegraphics[width=1.0\textwidth, trim={2.6cm, 0.4cm, 2.6cm, 0.5cm}, clip]{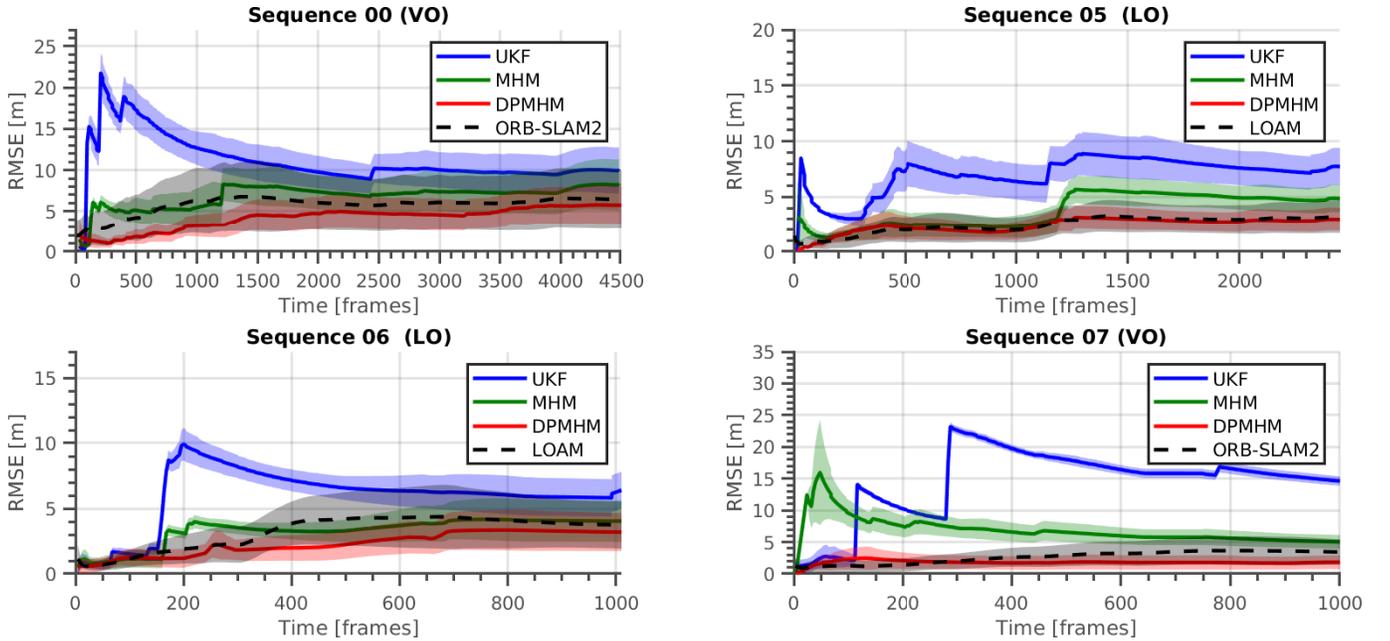}
   \caption{Comparison of several KITTI sequences by means of the \ac{RMSE} plotted as a function of the frames. The two methods, \ac{UKF} and MHM, show several jumps in the error due to wrong data associations. Each wrong association pulls the factor graph towards a wrong direction which results in the jumps of the \ac{RMSE}. Our method performs a more correct data association and keeps therefore the least error and does not include any sudden changes. }
   \label{pics:evaluation:rmse}
\end{figure*}
Every time a submap is completed, the resulting map as well as the odometry measurements are used to compute a trajectory estimate.
The weighted hypotheses allow for the creation of weighted mixtures of probability distributions resulting in a weighted fusion which considers the uncertainty of each hypothesis (cf. figure~\ref{pics:traj_with_GM}). 
The result of the fusion is formulated as a relative constraint and incorporated into a nonlinear factor graph as semantic landmarks.
\subsection{Semantic Evaluation of Submaps}
Loop closures are identified by first evaluating the quality of the submap in terms of the occurred landmarks. 
This is motivated by the fact that in many cases (e.g. highways) it is not necessary to check for loop closures.
The examination whether a submap is good enough for loop closure detection is based on a decision tree. 
We train a decision tree by comparing the trace of the state covariance before and after incorporating a specific region in the factor graph. 
The trained decision tree is specific to an urban environment and furthermore, to the length of the submap. 
Thus for other environments, a retraining of the decision tree or online learning approaches are required.
A submap is considered as good either when it lowers the size of the bounding box or by having loop closures in it. 

Evaluating a submap requires extracting descriptive attributes from it and we argue that semantic information is a crucial factor for this.
In more detail, we first approximate the Shannon entropy $H$ of the mixture distribution using an approximate single multivariate Gaussian distribution over the submap, with covariance $\mat{\Sigma}$, i.e.
\begin{align*}
      H 
      %
      %
      &= \frac{1}{2} \log \left( (2\pi{}e)^3 \det(\mat{\Sigma}) \right)\label{eq:normal:entropy}.
\end{align*}
On a level of semantic classes we then calculate a term frequency-inverse document frequency (tf-idf) score, i.e.
\begin{equation*}
S_{tf\text{-}idf}^i = \sum_c \frac{n_c^i}{n^i} \log\left(\frac{N}{n_c}\right),
\end{equation*}
where $n_c^i$ denotes the number of occurrences of class $c$ in submap $i$, $n^i$ the total number of classes in $i$. 
Furthermore, $N$ denotes the total number of submaps processed so far and $n_c$ represents the number of scenes within the submaps which included an object of type $c$. 
This is efficiently compared and updated with the previous submaps.
As a final score, we make use of the number of landmarks within the submap. 

As shown in figure \ref{pics:evaluation:maps}, the loop closure detection is triggered once the decision tree predicts that a submap is potentially good in terms of its mapped objects.
\subsection{Semantic Loop Closure Detection}
Loop closures are found in multiple steps. First, we find similar submaps using an incremental kd-tree~\cite{Bentley1975} of the submap's normalized class histograms while  
%
employing the Jensen-Shannon divergence (JSD)~\cite{Lin1991} as the distance measure.
For each similar submap the individual scene candidates are identified with another kd-tree of the scene's normalized class histograms using the $L_2$-norm for faster retrieval.
Additional efficiency can be achieved with a tuning parameter that restricts the search space of the kd-tree in terms of the distance.
Good loop closure candidates found by the second kd-tree are verified and further filtered with a discrete Bayes filter.
We define a Markov chain between the events for loop closure and no loop closure. 
The transitional probabilities are chosen to be similar to~\cite{Angeli2008}.

The verification process calculates two scores for how similar the candidate scene and the current scene are. 
First, the topology of a scene is represented by the Laplacian matrix which is calculated based on the spatial relationship between the semantic classes as well as their degrees in the scene.
We compare the topologies of two scenes based on a normalized cross correlation (NCC)~\cite{Cascianelli2017} score, $S_{NCC}$.
Second, another score, $S_{scene}$, expresses the overall similarity of the landmarks in the two scenes.
For this the landmarks get associated with the Hungarian algorithm on the estimated landmark positions of each scene and their Euclidean distances. 
A pair of matched landmarks $i$ and $j$ contribute to $S_{scene}$ through
\begin{equation*}
  s_{match}^{i,j} := 1 - \frac{\mat{H}_{i,j}}{2}, \hspace{0.8cm} s_{class}^{i,j} := (1-\delta_{c_i,c_j}) \cdot p,
\end{equation*}
where $\mat{H}$ is the Hungarian cost matrix (output of the Hungarian algorithm), $p$ denoting a penalty factor, $c_i$, $c_j$ being the label of the landmarks $i,j$, and $\delta$ denoting the Kronecker delta. 
The two scores $s_{match}$ and $s_{class}$ are combined using all matched landmark pairs, as follows
\begin{equation*}
  S_{scene} := \sum_{i,j} 1 - s_{match}^{i,j} s_{class}^{i,j}.
\end{equation*}
The sum of both scores, $S_{NCC}$ as well as $S_{scene}$ has to be larger than a threshold (tuning parameter) to verify the match of the two scenes. 
This binary decision serves as input to the discrete Bayes filter which finally gets to decide whether to use the scene pair as a loop closure candidate for the next step. 
As the last step, the set of all loop closures candidates undergoes a final geometric consistency check based on RANSAC before the actual loop closure constraints are inserted into the factor graph. 
Both, the Hungarian algorithm and RANSAC can be computationally expensive.
Therefore, we filter most invalid candidates beforehand using the kd-trees which can be performed in logarithmic time.
For additional robustness, we use m-estimators with Cauchy functions~\cite{lee2013robust} in the optimization of the factor graph.

\section{Evaluation}
\label{sec:evaluation}
We evaluate our system on the KITTI dataset sequences 00, 05, 06 and 07~\cite{Geiger2012} where we use SegNet~\cite{Badrinarayanan2017} to derive the semantic classes of the individual scenes.
For each image, the semantic objects are extracted and projected into the world frame~\cite{Gawel2017a} using the Velodyne scans.
In general, our approach is not limited to the use outdoors but rather depends on the object detector.
Additionally, one might need to adapt the decision tree and $p_s$ in equation~\eqref{eq:meas_likelihood}.
Since, to our knowledge, no appropriate approach for comparison is publicly available, we could not compare our proposed approach to another semantic SLAM system. 
Therefore, we evaluate our proposed system to two other semantic solutions as well as two non-semantic approaches.
As a baseline we compare to LeGo-LOAM\cite{shan2018lego} and ORB-SLAM2 stereo~\cite{Mur-Artal2017}. 
For a semantic baseline, we utilize a single UKF estimator with a Hungarian algorithm based on the $L_2$ norm for data association. 
This essentially performs a nearest neighbor data association with a single hypothesis. 
We also added a multiple hypothesis mapping (MHM) using a maximum likelihood approach and a \ac{MHt}.
Similar to our main approach, frequent optimizations of the \ac{MHt} are needed. 
Hence, we threshold the likelihood if the \ac{MHt} reaches a certain size (see equation \eqref{eq:meas_likelihood}) and keep only the best third of all. 

Our proposed system is agnostic to the source of odometry which we show by making use of two different ones for all sequences. 
More specifically, we utilized the tracking of ORB features~\cite{Mur-Artal2017} as well as LiDAR surface and corner features~\cite{shan2018lego} to get an odometry estimate.
We have used the provided camera calibration parameters from Geiger et al.~\cite{Geiger2012} for both, \ac{VO} and ORB-SLAM2.
Thus, the results of ORB-SLAM2 are different than the results reported in the work of Mur-Artal et al.~\cite{Mur-Artal2017} where they used different parameters per sequence.
\subsection{Results}
We demonstrate the performance of our proposed approach by means of calculating the \ac{RMSE} of the estimated trajectory location to the GPS ground truth provided by the KITTI dataset using the \ac{VO} and \ac{LO} sources. 
Both, \ac{VO} and \ac{LO}, accumulate an error and hence, are subject to drift over time. 
Using our \ac{DP}-based multiple hypothesis mapping approach together with our place recognition (cf. figure \ref{pics:evaluation:maps}) we can reduce the drift up to 50\,\% for several sequences. 

\begin{figure}[!htb]
   \centering
   \includegraphics[width=0.48\textwidth, trim={0.3cm, 0.2cm, 0.3cm, 1.6cm}, clip]{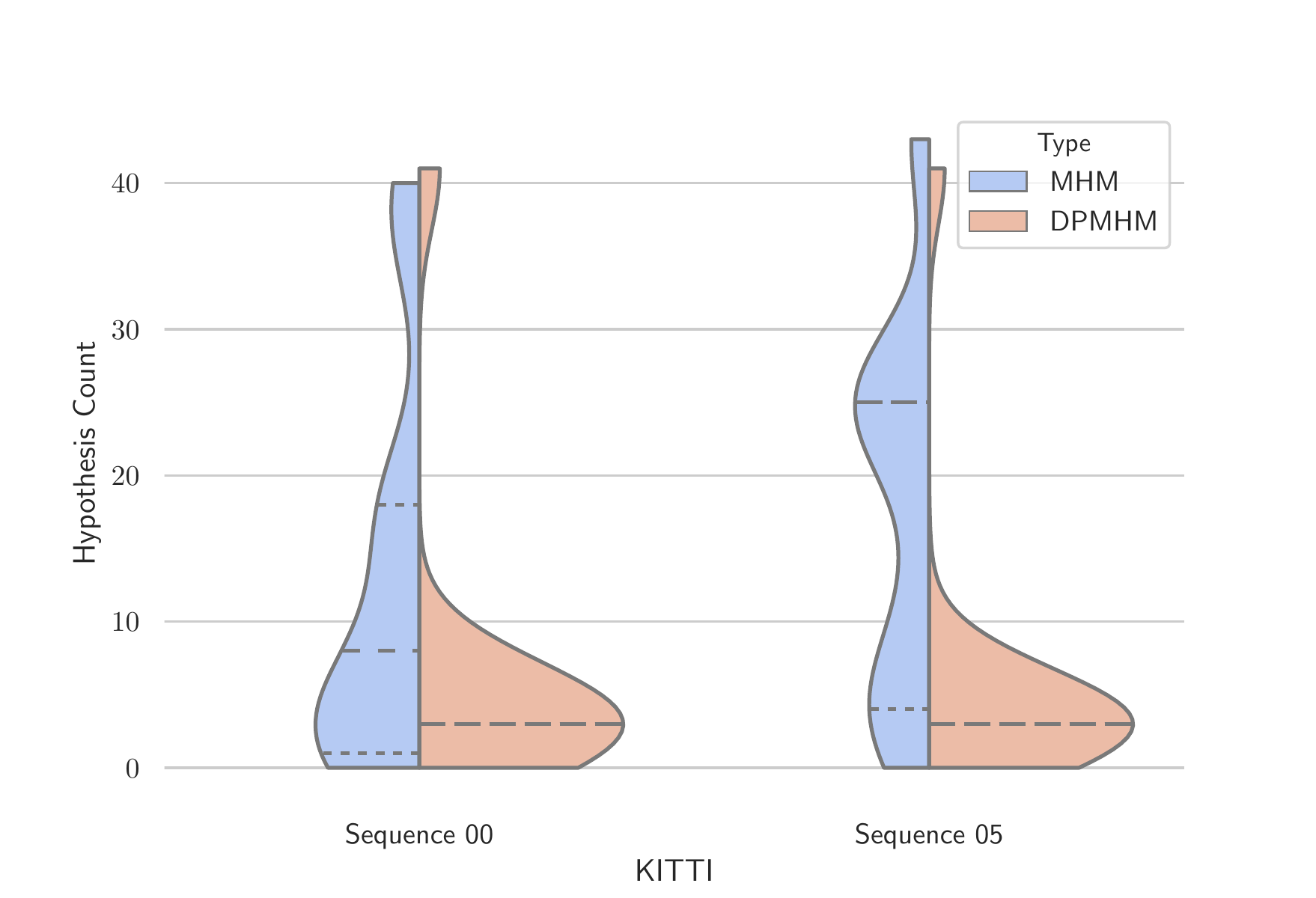}
   \caption{Evaluation of the two multiple hypotheses-based implementations. The naive likelihood thresholding approach has an average of 12 (sequence 00) and 18 (sequence 05) hypotheses, respectively, whereas our proposed resampling approach has an average of 7 (sequence 00) and 6 (sequence 05) hypotheses.}
   \label{pics:violin_hypotheses_count}
\end{figure}

\begin{figure}[!htb]
   \centering
   \includegraphics[width=0.48\textwidth, trim={2.8cm, 12.5cm, 3.4cm, 11.303cm}, clip]{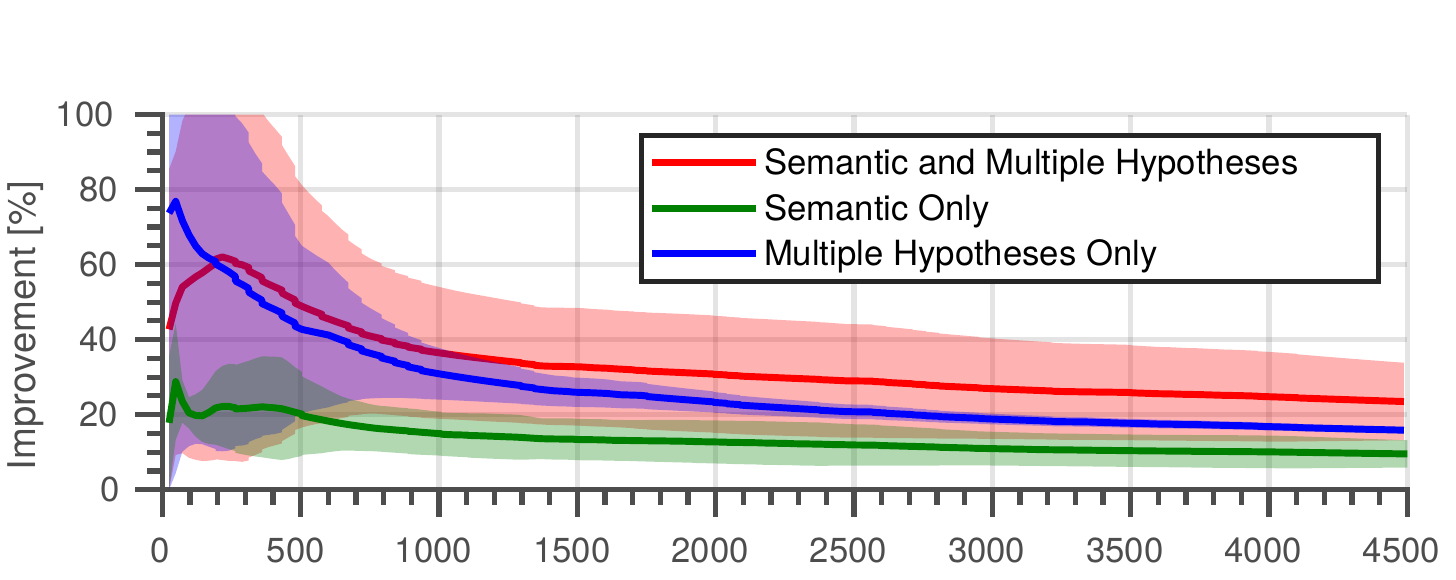}
   \caption{Reduction of the RMSE for a single hypothesis, non-semantic DPMHM. Including both modalities, we achieve an average reduction of 34\,\%, with only multiple hypotheses 27\,\% and 13\,\%  with a pure semantic DPMHM. }
   \label{pics:improvement}
\end{figure}

The simple UKF and MHM estimation approaches are strongly affected by wrong measurement associations resulting in bad constraints in the pose graph.  
These wrong assignments can be observed in figure \ref{pics:evaluation:rmse} as sudden jumps in the \ac{RMSE}. 
Consequently, the \ac{RMSE} will have an increased total error which is even worse than the raw odometry source for a few sequences. 
Our approach is less perturbed with wrong associations and thus, maintains a more robust \ac{RMSE} over time.
Table~\ref{table:results:statsKittiVO} show the mean and standard deviation of the \ac{RMSE} for each sequence and estimator.
Our approach does particularly well on on the longer sequences (00, 05) which results from the correct data association together with the semantic place recognition. 
Regardless of the odometry source, our proposed system yields results comparable to the state-of-the-art SLAM approaches in \ac{VO} and \ac{LO} and compared to the MHM, maintains less hypotheses about the environment as shown in figure~\ref{pics:violin_hypotheses_count}. 
Due to the fact that the total number of hypotheses of the environment is only increased when the state uncertainty is high, we gain additional efficiency for our proposed system.
Figure~\ref{pics:improvement} evaluates the performance of our algorithm when the semantic classes are removed as well as for the restriction to a single hypothesis.
The single hypothesis, no semantic solution then still performs a probabilistic Hungarian method and achieves a mean RMSE of \SI{5.45}{\metre}$\pm$\SI{2.94}{\metre}.
\begin{table}[!htb]
\vspace{0.3cm}
\begin{center}
    \begin{tabular}{ *5c }
    \toprule
    Sequence & 00 & 05 & 06 & 07\\
    \midrule
     VO & 8.41$\pm$2.51 & 6.42$\pm$3.9 & 3.8$\pm$1.4 & 6.23$\pm$2.4  \\
     UKF & 11.14$\pm$2.9 & 8.9$\pm$4.2 & 5.96$\pm$2.4  & 14.55$\pm$5.1  \\
     MHM & 6.84$\pm$1.3  & 5.6$\pm$2.94 & 3.17$\pm$1.07 & 6.93$\pm$2.02 \\
     DPMHM & \textbf{4.54}$\pm$1.58  & \textbf{4.4}$\pm$2.3 & 2.3$\pm$0.74&2.9$\pm$4.5 \\
     ORB-SLAM2 & 5.7$\pm$\textbf{1.0} & 4.51$\pm$\textbf{1.3} & \textbf{2.1}$\pm$\textbf{0.6} & \textbf{2.71}$\pm$\textbf{0.9}  \\
     \midrule
     LO & 7.33$\pm$2.5 & 2.96$\pm$1.3 & 3.3$\pm$1.2 & 6.65$\pm$2.85  \\
     UKF & 6.96$\pm$1.7 & 6.96$\pm$1.5 & 5.47$\pm$0.75  & 10.4$\pm$2.8  \\
     MHM & 5.3$\pm$2.2  & 3.8$\pm$1.4 & 2.67$\pm$\textbf{0.35} & 11.3$\pm$3.9 \\
     DPMHM & \textbf{3.94}$\pm$\textbf{1.17}  & \textbf{2.42}$\pm$\textbf{0.66} & 2.66$\pm$\textbf{0.35}& 5.5$\pm$2.4 \\
     LeGo-LOAM & 5.8$\pm$2.2 & 2.54$\pm$0.72 & \textbf{2.15}$\pm$0.52 & \textbf{1.0}$\pm$\textbf{0.16} \\
    \bottomrule
    \end{tabular}
    \caption{Comparison of the mean RMSE and standard deviation in meters achieved with \ac{VO} and \ac{LO} as the underlying odometry source.}
    \label{table:results:statsKittiVO} 
\end{center}
\end{table} 
\section{Conclusion and Future Work}
\label{sec:conclusion}

In this work, we presented a novel semantic SLAM system based on factor graphs and a \ac{MHt} mapping approach aiming to deal with ambiguities in data association in semantic-based SLAM.
We showed that our resampling method for optimizing the hypothesis tree yields a more robust estimation and requires substantially less hypotheses. Moreover, we gain additional efficiency by preselecting submaps for loop closure detection.

As further research, we intend to remove the assumption that each object can generate at most one measurement per time-step since a bad detector or viewpoint angle might easily violate this assumption. Additionally, this work could potentially also be extended towards utilizing an instance-based detection. Instance information could possibly give a prior on how to associate the measurements at the cost of an additional non-Gaussian discrete random variable.




\bibliographystyle{IEEEtran_Capitalize}
\bibliography{bibtex/lb,bibtex/libraries}

\end{document}